\documentclass[conference]{IEEEtran}

\IEEEoverridecommandlockouts                              

\pdfoutput=1

\usepackage{amsmath,amsthm,amssymb} 
\usepackage{style/ko_yamamoto}
\usepackage{graphicx}
\usepackage{color,soul}
\usepackage{booktabs}
\usepackage{changes}
\usepackage[ruled,linesnumbered]{algorithm2e}  
\usepackage{algorithmic}
\usepackage{subfig}
\usepackage{threeparttable}

\usepackage[normalem]{ulem}
\def\modified_color{black}
\def\rev_color{black}

\usepackage{amsthm}
\newtheorem{proposition}{Proposition}

\theoremstyle{definition}
\newtheorem{definition}{Definition}
\usepackage{hyperref}

\usepackage{caption}
\title{
Task-Specified Compliance Bounds for Humanoids via Lipschitz-Constrained Policies
}

\author{Zewen He and Yoshihiko Nakamura 
\thanks{The authors are with the Department of Robotics,
Mohamed bin Zayed University of Artificial Intelligence, Masdar City, Abu Dhabi, United Arab Emirates. 
        {\tt\small \{zewen.he, yoshihiko.nakamura\}@mbzuai.ac.ae}}%
}

\begin{document}

\maketitle
\thispagestyle{empty}
\pagestyle{empty}

\begin{abstract}
Reinforcement learning (RL) has demonstrated substantial potential for humanoid bipedal locomotion and the control of complex motions.
To cope with oscillations and impacts induced by environmental interactions, compliant control is widely regarded as an effective remedy.
However, the model-free nature of RL makes it difficult to impose \emph{task-specified} and quantitatively verifiable compliance objectives, and classical model-based stiffness designs are not directly applicable.
Lipschitz-Constrained Policies (LCP), which regularize the local sensitivity of a policy via gradient penalties, have recently been used to smooth humanoid motions.
Nevertheless, existing LCP-based methods typically employ a single scalar Lipschitz budget and lack an explicit connection to physically meaningful compliance specifications in real-world systems.
In this study, we propose an anisotropic Lipschitz-constrained policy (ALCP) that maps a task-space stiffness upper bound to a
state-dependent Lipschitz-style constraint on the policy Jacobian.
The resulting constraint is enforced during RL training via a hinge-squared spectral-norm penalty,
preserving physical interpretability while enabling direction-dependent compliance.
Experiments on humanoid robots show that ALCP improves locomotion stability and impact robustness, while reducing oscillations and energy usage.

\end{abstract}

\section{Introduction\label{sect:introduction}}

\color{black}

Humanoid robots, owing to their human-like morphology, have long been regarded as a general-purpose form factor for enabling robots to truly integrate into human society.
Since their inception, a core challenge for humanoid robots has been maintaining stability during interactions with the environment. 
On the one hand, humanoids inherently rely on bipedal locomotion, which is intrinsically less stable than fixed-base mechanisms; 
on the other hand, in practical applications, robots must physically interact with the environment and, potentially, with humans.
Research efforts—from early simplified models \cite{kajita2003biped}, through whole-body model-based control (e.g., model predictive control) \cite{escande2014hierarchical}, to reinforcement learning (RL) policies \cite{radosavovic2024real}—have consistently revolved around this fundamental challenge.

Although current RL-based control policies can satisfy the basic requirements of bipedal locomotion, their black-box nature makes it difficult to reliably enforce many specific and quantifiable objectives beyond ``walk without falling" \cite{radosavovic2024real}. 
A representative example is task-specified compliant control, which is crucial for task performance and safety when humanoid robots execute practical manipulation or physical-interaction tasks. 
Using foot impact during stepping as an example, many RL locomotion pipelines mitigate undesired transients by adding shaping penalties (e.g., on joint torques, accelerations, or action-rate changes), which can reduce impact-related behaviors \cite{rudin2022learning}.
While often effective, this reward-engineering strategy does not naturally yield an interpretable, analyzable compliance specification, and it can introduce objective coupling: the same penalty weights that improve smoothness or reduce impacts may also hinder task performance, requiring tedious retuning across domains \cite{chen2025learning}. 

\begin{figure}[t]
  \centering
  \subfloat[]{
    \includegraphics[width=.39\linewidth]{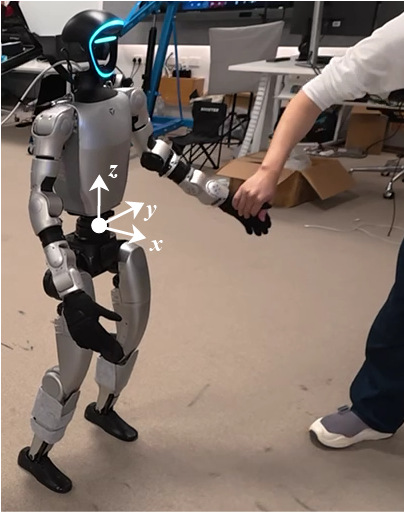}
    \label{fig:g1_exp_a}
  }\hspace{-0.5mm}
  \subfloat[]{
    \includegraphics[width=.22\linewidth]{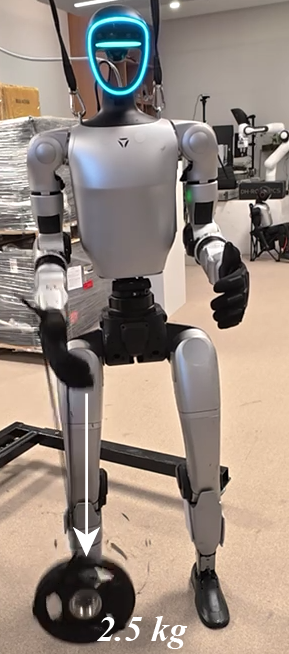}
    \label{fig:g1_exp_b}
  }\hspace{-0.5mm}
  \subfloat[]{
    \includegraphics[width=.25\linewidth]{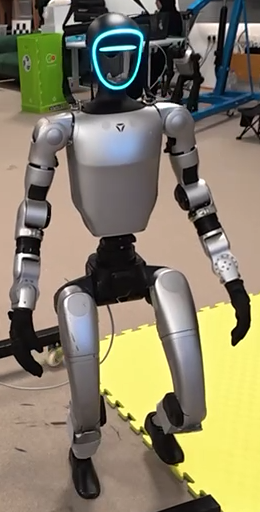}
    \label{fig:g1_exp_c}
  }
  \caption{{Task-specific compliance. }(a) Hand-pulling experiment under the SILC method. (b) Hand impact experiment. A 2.5 kg weight is attached to the robot's right hand via a rope and released from a certain height to generate an impact. (c) Stepping motion experiment. The robot's one foot intermittently steps onto a foam pad placed on the ground, thereby introducing disturbances from uneven terrain. 
}
  \label{fig:g1_exp}
\end{figure}

To address the same issue, a broad range of approaches has been developed in classical model-based robot control \cite{henze2015approach,yamamoto2021humanoid}.
However, most of these methods rely purely on model-based theory and require sufficient hardware support (e.g., force sensors). 
For a long time, these methods and learning-based approaches progressed largely in parallel, with little connection between them, and thus cannot simultaneously benefit from the advantages offered by RL-based approaches.
Motivated by this observation, recent studies have attempted to achieve compliant control within RL-based control frameworks.
\sectref{sec:compliant} of this paper reviews these efforts.
Overall, however, most existing studies effectively wrap model-based controllers around the RL framework, without explicitly observing the equivalent stiffness or analyzing the internal control behavior induced by the policy itself.

Meanwhile, in RL-based methods, the use of simplified dynamics and actuation models often biases the policy toward jittery behaviors, which can severely compromise stability in real-world experiments \cite{chen2025learning}.
To mitigate this phenomenon, in addition to directly adding smoothness rewards \cite{he2024learning} or a low-pass filter \cite{li2021reinforcement}, introducing Lipschitz constraints to the policy training is considered a more effective approach, which is referred to as Lipschitz-constrained policies (LCP) \cite{chen2025learning,shin2025spectral}.
However, in existing LCP methods applied to humanoid robots, the Lipschitz constant is typically set as a scalar. 
Moreover, in practical training, it is relaxed into a soft constraint or enforced only implicitly, which undermines its physical interpretability.

Based on the above state of the art, this paper aims to start from quantitatively specified task compliance and, based on Lipschitz-constraint theory, enforce an equivalent stiffness constraint within an RL framework on humanoids. 
We propose a \textbf{Stiffness-Induced Lipschitz Constrained (SILC)} policy, which maps a prescribed stiffness upper-bound (in task or joint space) to an anisotropic Lipschitz budget on the policy Jacobian, and enforces it via {a LCP-style regularization. In this study, we use \textit{Lipschitz-style regularization} to denote Jacobian-norm penalties that cap a policy's local sensitivity}. 
Specifically, we first use kinematic relations to map the task-space stiffness constraint to a corresponding joint-space stiffness budget. We then combine this budget with the low-level control law to express it as a target Lipschitz-style constraint, which is incorporated into the loss function during RL training.
Finally, in experiments (as shown in \figref{fig:g1_exp}), by specifying different target task-space stiffness values, we achieve tunable compliance during whole-body learning-based control on a humanoid robot.

The main contributions of this study are as follows:
\begin{enumerate}
    \item By leveraging the Jacobian of the RL policy, we propose \textbf{policy-induced equivalent joint stiffness}, which enables explicit observation of the system’s equivalent joint stiffness under an RL-based control framework;
    \item We extend scalar Lipschitz-constrained policies to \textbf{anisotropic Lipschitz-constrained policies (ALCP)} for RL training, and derive target ALCP budgets from task-space stiffness constraints, thereby endowing ALCP with a clear physical interpretation for the robotic system;
    \item Experimental results demonstrate that the proposed ALCP enables both task-specified compliance and stable humanoid locomotion.
\end{enumerate}



\section{Background\label{sect:background}}




\subsection{Learning-based Humanoid Control}

In recent years, RL-based motion control for legged robots—particularly humanoid robots—has largely converged toward a well-established paradigm. 
Through appropriate reward design in RL, a humanoid robot’s basic stable gait can generally be ensured \cite{guevolution}. 
Furthermore, researchers have employed imitation-learning approaches (e.g., AMP \cite{peng2021amp}) to enable robots to mimic human motions while still maintaining their own stability. 
Meanwhile, research on humanoid robots has shifted from pure locomotion control toward loco-manipulation \cite{gu2025humanoid}.
Whether in conventional locomotion control or more complex loco-manipulation, the central challenges for humanoid robots remain unchanged: ensuring balance under a floating-base formulation and handling physically reasonable contacts with the environment.
To address these issues within learning-based control, many classical model-based control components have been integrated into the framework, such as compliance control.

\subsection{Compliant Control}
\label{sec:compliant}
Classical model-based compliant control has been studied for many years, and it has also achieved substantial progress in the humanoid robotics community \cite{henze2015approach}.
Recently, there has been a growing interest in integrating classical compliance control into RL-based humanoid robot control.
A key challenge in this integration is how to preserve the physical interpretability of compliance while retaining the distinctive advantages of RL-based methods.
In early work such as deep compliant control \cite{lee2022deep}, compliance objectives were recast as admittance-like tracking targets and learned via imitation. This ``admittance-to-reference" paradigm was later extended in \cite{xu2025facet,zhao2025resmimic}, where desired compliance is mapped to a target displacement (in task or joint space) and the policy is trained to track the resulting reference. In parallel, other studies embed model-based compliant control directly into RL, producing behavior closer to classical impedance control \cite{wei2025hmc,he2025cotap}. 
The former integrates cleanly with RL but can blur the intended physical meaning of compliance under learning-based tracking, whereas the latter is physically intuitive yet may conflict with the learned policy (e.g., overly stiff low-level loops can induce oscillations). 
Finally, neither of the above two categories provides a way to observe the system's overall compliance characteristics, which makes targeted compliance control considerably inconvenient. 

\subsection{Lipschitz Contunuity and Lipschitz-constrained Policies}
\label{sec:lip_constr}
\subsubsection{Lipschitz continuity}
Let $(\mathcal{X},\|\cdot\|_{\mathcal{X}})$ and $(\mathcal{Y},\|\cdot\|_{\mathcal{Y}})$ be normed spaces and
$f:\mathcal{X}\rightarrow\mathcal{Y}$ be a function. We say that $f$ is \emph{Lipschitz continuous} if there exists a
constant $L\ge 0$ such that, for all $\mathbf{x}_1,\mathbf{x}_2\in\mathcal{X}$,
\begin{equation}
\|f(\mathbf{x}_1)-f(\mathbf{x}_2)\|_{\mathcal{Y}}
\le
L\,\|\mathbf{x}_1-\mathbf{x}_2\|_{\mathcal{X}}.
\end{equation}
The smallest such $L$ is called the \emph{Lipschitz constant} of $f$, denoted by $\mathrm{Lip}(f)$.
If $f$ is differentiable, a common sufficient bound is
\begin{equation}
\mathrm{Lip}(f)
\ge
\sup_{\mathbf{x}\in\mathcal{X}}
\left\|\frac{\partial f(\mathbf{x})}{\partial \mathbf{x}}\right\|_{2},
\end{equation}
where $\|\cdot\|_{2}$ denotes the spectral (operator) norm.

\subsubsection{Lipschitz-Constrained policies (LCP)}
In reinforcement learning, a (deterministic) policy $\pi_{\boldsymbol{\theta}}:\mathcal{O}\rightarrow\mathcal{A}$ maps an
observation $\mathbf{o}\in\mathcal{O}$ to an action $\mathbf{a}\in\mathcal{A}$.
A policy is called \emph{Lipschitz-constrained} if its input--output sensitivity is bounded, i.e., there exists
$K_{\mathrm{LCP}}\ge 0$ such that
\begin{equation}
\|\pi_{\boldsymbol{\theta}}(\mathbf{o}_1)-\pi_{\boldsymbol{\theta}}(\mathbf{o}_2)\|_{2}
\le
K_{\mathrm{LCP}}\,
\|\mathbf{o}_1-\mathbf{o}_2\|_{2},
\quad \forall\,\mathbf{o}_1,\mathbf{o}_2\in\mathcal{O}.
\end{equation}
Equivalently, for differentiable policies, one can enforce a local Jacobian bound,
\begin{equation}
\left\|\frac{\partial \pi_{\boldsymbol{\theta}}(\mathbf{o})}{\partial \mathbf{o}}\right\|_{2}
\le
K_{\mathrm{LCP}},
\quad \forall\,\mathbf{o}\in\mathcal{O},
\label{eq:klcp}
\end{equation}
or relax it using sample-based penalties during training.
For stochastic policies $\pi_{\boldsymbol{\theta}}(\mathbf{a}\mid \mathbf{o})$, Lipschitz constraints are often imposed via regularizing the norm of the input-gradient of the log-likelihood, $\|\nabla_{\mathbf{o}}\log \pi_{\boldsymbol{\theta}}(\mathbf{a}\mid \mathbf{o})\|$.
In several prior studies, LCP-based methods have been incorporated into RL training for humanoid robots; however, their primary objective is to mitigate motion oscillations, and they typically consider only an isotropic Lipschitz constraint \cite{chen2025learning,shin2025spectral}.

\begin{figure}[t]
	\begin{center}
	 \includegraphics[width=0.98\hsize]{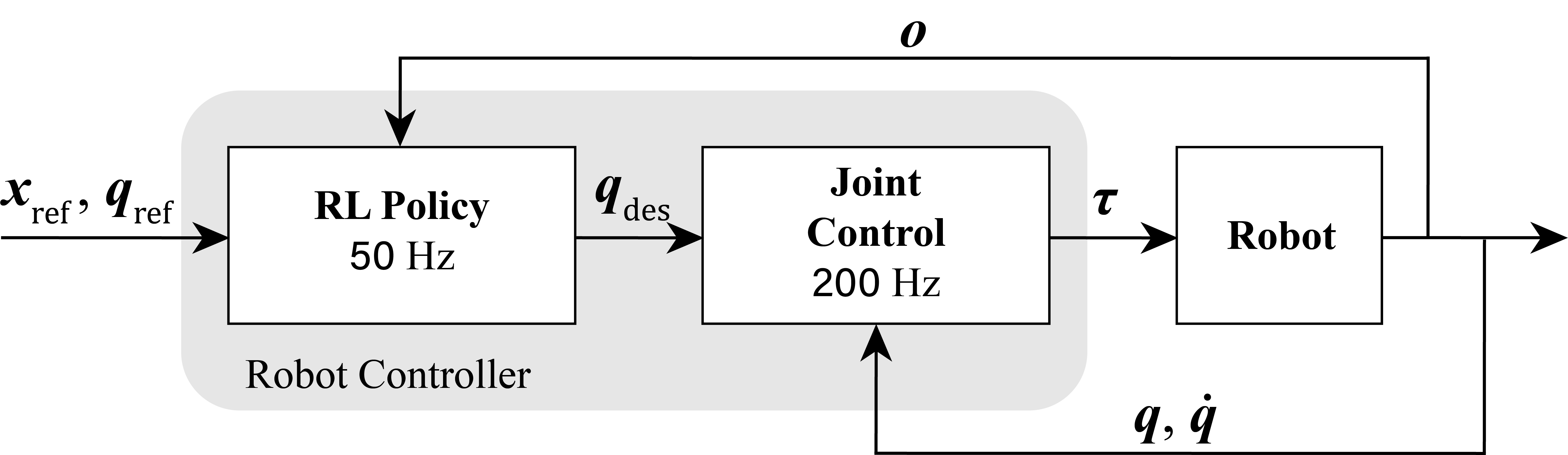}
	 \caption{Control framework in reinforcement learning pipeline. }
    \label{fig:ctrl_frame}
	\end{center}
\end{figure}

\section{Problem Statement}
\label{sec:problem_statement}

We consider a legged humanoid robot controlled by a hierarchical architecture. The low-level controller is a fixed joint-space PD
position servo, while a high-level RL policy outputs joint position targets.
Let $\bm q\in\mathbb{R}^{n}$ and $\dot{\bm q}\in\mathbb{R}^{n}$ denote joint positions and velocities, and let
$\bm o\in\mathcal{O}$ be the policy observation (e.g., full proprioception). The policy
$\bm q_{\mathrm{des}}=\pi_{\theta}(\bm o)$ produces desired joint positions, and the PD controller applies joint torques
\begin{equation}
\bm\tau = \bm K_p(\bm q_{\mathrm{des}}-\bm q) - \bm K_d \dot{\bm q},
\label{eq:pd_controller}
\end{equation}
where $\bm K_p\succ\bm 0$ and $\bm K_d\succeq \bm 0$ are constant gain matrices (typically diagonal). 
This RL-based control framework can be schematically illustrated in \figref{fig:ctrl_frame}.

Let $\bm x\in\mathbb{R}^{m}$ denote a task-space coordinate of interest (e.g., center of mass (CoM) position), with Jacobian $\bm J_x(\bm q)\triangleq \partial \bm x/\partial \bm q$.
In the neighborhood of a nominal motion, a task-space impedance law is commonly written as
\begin{equation}
\bm f_x
=
\bm K_x\big(\bm x_{\mathrm{des}}-\bm x\big) - \bm D_x \dot{\bm x},
\label{eq:task_pd}
\end{equation}
where $\bm f_x\in\mathbb{R}^{m}$ is the {task-space force}, and $\bm K_x\succeq \bm 0$, $\bm D_x\succeq \bm 0$
denote the task-space stiffness and damping, respectively. 
{Here we consider that $\bm K_x$ and $\bm D_x$ are always symmetric positive semidefinite (SPD). }
The task-space stiffness matrix $\bm K_x$ defines a \emph{stiffness ellipsoid} {with a positive constant $c$} through the quadratic form
\begin{equation}
\mathcal{E}_x(\bm K_x, c)
\triangleq
\left\{
\delta\bm x\in\mathbb{R}^{m}:\ \delta\bm x^{\top}\bm K_x\,\delta\bm x \le c
\right\},
\label{eq:task_stiff_ellipsoid}
\end{equation}
We are given a task-level SPD stiffness budget $\bm K_x^{\max}$ and require
$
\mathcal{E}_x(\bm K_x, c)\ \subseteq\ \mathcal{E}_x(\bm K_x^{\max}, c)
\label{eq:ellipsoid_inclusion}
$, which means that the ellipsoid induced by $\bm K_x^{\max}$ upper-bounds the allowable stiffness in every direction. 
In this work, $\bm K_x^{\max}$ is set as a diagonal matrix. 
For a position task, $\bm K_x^{\max}$ can be visualized as a spatial ellipsoid.

We aim to learn parameters $\bm\theta$ that maximize the expected return under an RL objective while enforcing (or approximately
enforcing) a Lipschitz constraint on the policy sensitivity:
\begin{equation}
\max_{\bm\theta}\ \mathbb{E}\!\left[\sum_{t=0}^{T} \gamma^t r(\bm s_t,\bm a_t)\right]
\quad
\text{s.t.}\ 
\left\|
\frac{\partial \pi_{\theta}(\bm o_t)}{\partial \bm o_t}
\right\|_{2}\le
K_{\mathrm{LCP}},
\ \forall t,
\label{eq:rl_constrained_problem}
\end{equation}
where $\bm a_t=\pi_{\theta}(\bm o_t)$ and the constraint may be enforced either as a hard constraint or via a soft penalty during
training.

\begin{figure}[t]
	\begin{center}
	 \includegraphics[width=0.95\hsize]{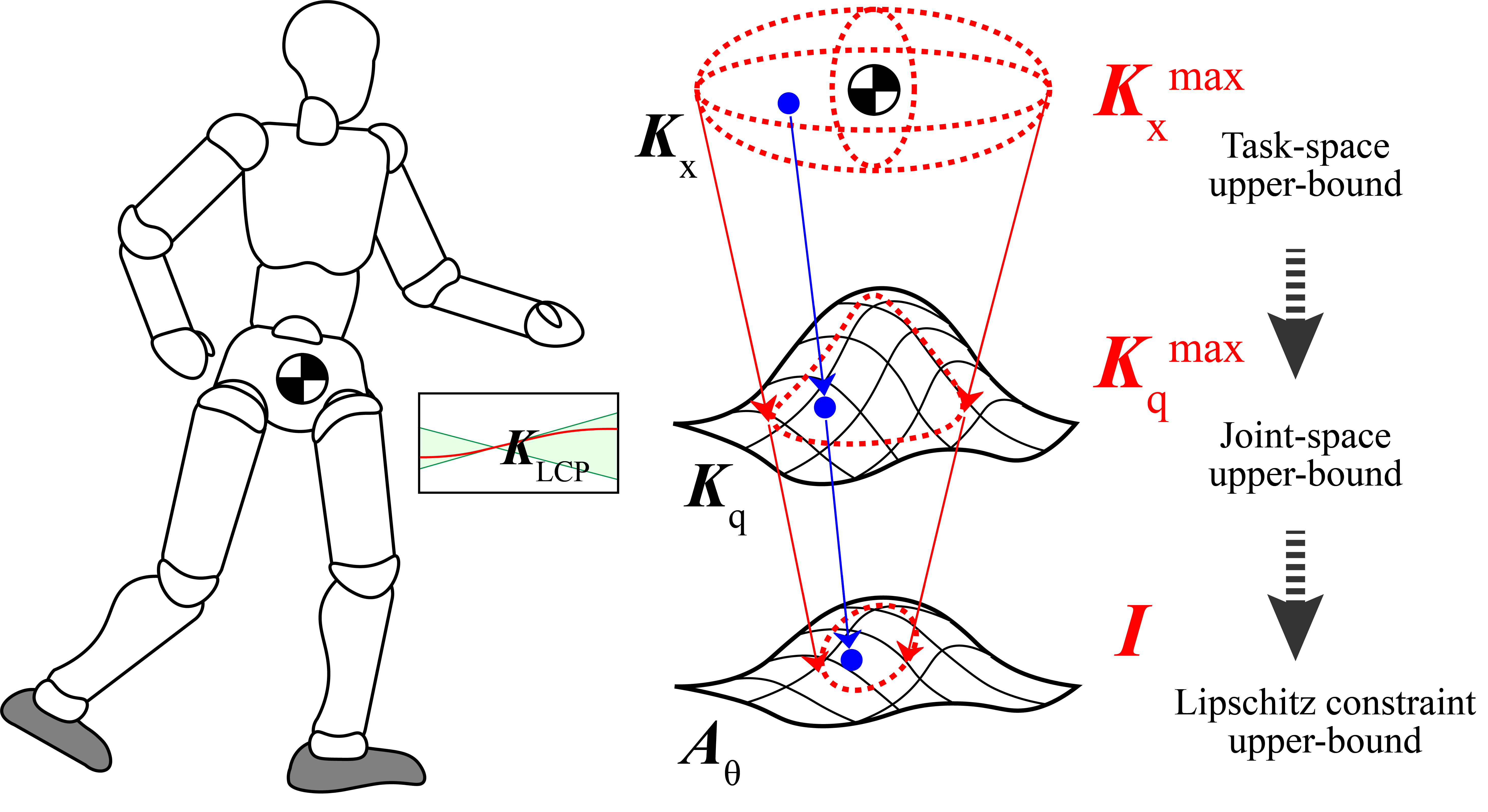}
	 \caption{Stiffness upper-bound derivation diagram. $\bm K_x^{\max}$ denotes the task-space stiffness upper-bound ellipsoid, illustrated as the pink ellipsoid; $\bm K_q^{\max}$ denotes the joint-space stiffness upper-bound super-ellipsoid, where individual joint stiffnesses are depicted as torsional springs; $\bm K_{LCP}$ represents the Lipschitz upper-bound constraint. }
    \label{fig:stiff_diagram}
	\end{center}
\end{figure}

\begin{proposition}[\textbf{Policy-induced equivalent joint stiffness}]
\label{prop:policy_induced_stiffness}
Assume that the observation $\bm o$ contains the joint positions $\bm q$ and that $\pi_{\theta}$ is differentiable with respect
to $\bm q$ in a neighborhood of interest. Then, for small perturbations around a nominal state, the incremental torque response to
joint-position perturbations satisfies
\begin{equation}
\delta\bm\tau
=
-\bm K_{\mathrm{eq}}(\bm o)\,\delta\bm q
\;+\;
\mathcal{O}\!\left(\big\|[\delta\bm q;\delta\dot{\bm q}]\big\|^2\right),
\label{eq:delta_tau_linearization}
\end{equation}
where the \emph{policy-induced equivalent joint stiffness} is given by
\begin{equation}
\bm K_{\mathrm{eq}}(\bm o)
\triangleq
-\frac{\partial \bm\tau}{\partial \bm q}
\approx
\bm K_p\!\left(\bm I - \alpha \bm J_\pi(\bm o)\right),
\quad
\bm J_\pi(\bm o)\triangleq \frac{\partial \pi_{\theta}(\bm o)}{\partial \bm q}.
\label{eq:Kq_equiv_prop}
\end{equation}
where $\alpha$ is the action scale of the policy. 
{The derivation of $\bm K_{\mathrm{eq}}$ is given by \eqref{eq:pd_controller} and $\bm q_{\mathrm{des}} = \alpha \bm a_t $. }
\end{proposition}
Based on the above derivations, since it accounts for both the prescribed low-level stiffness and the policy’s intrinsic sensitivity, $\bm K_{\mathrm{eq}}$ provides a more comprehensive measure for observing compliance.

Given a desired task-space stiffness upper bound $\bm K_x^{\max}$, our goal is to train a policy whose LCP-regularized local sensitivity induces an effective joint stiffness $\bm K_{\mathrm{eq}}(\bm o)$ consistent with the task-level specification. 
Formally, letting $\bm K_x(\bm o)$ denote the task-space stiffness induced by $\bm K_{\mathrm{eq}}(\bm o)$, we seek a training procedure such that
\begin{equation}
\bm K_x(\bm o)
\preceq
\bm K_x^{\max},
\label{eq:task_ellipsoid_constraint_goal}
\end{equation}
by appropriately choosing anisotropic Lipschitz budgets in the policy regularization. 
Following the above derivations, the procedure used in this paper to infer the target upper-bound of the LCP from the task-space stiffness is illustrated in \figref{fig:stiff_diagram}.

\section{Methodology}

\subsection{Anisotropic Lipschitz-Constrained Policies}
\label{sec:matrix_KLCP_to_loss}

\subsubsection{Anisotropic LCP upper-bound}

Building on the scalar upper-bound LCP introduced in \sectref{sec:lip_constr}, we extend it to the anisotropic setting in this study. Accordingly, we define the following.
\begin{definition}[\textbf{Anisotropic Lipschitz constraint}]
\label{def:anisotropic_lip}
Let $\pi_{\theta}:\mathcal{O}\to\mathbb{R}^n$ be differentiable in $\bm o$, with policy Jacobian
$\bm J_\pi(\bm o)$.
Given a non-singular matrix $\bm K_{\mathrm{LCP}}\in\mathbb{R}^{n\times n}$, we say that $\pi_{\theta}$ satisfies an
\emph{anisotropic Lipschitz constraint} with budget $\bm K_{\mathrm{LCP}}$ on $\mathcal{O}$ if
\begin{equation}
\bm J_\pi(\bm o)\bm J_\pi(\bm o)^{\top}
\ \preceq\
\bm K_{\mathrm{LCP}}\bm K_{\mathrm{LCP}}^{\top},
\qquad
\forall\,\bm o\in\mathcal{O}.
\label{eq:hard_matrix_constraint_def}
\end{equation}
\end{definition}



Definition~\ref{def:anisotropic_lip} specifies a \emph{matrix} (anisotropic) generalization of the standard scalar Lipschitz bound in \eqref{eq:klcp}.
Geometrically, $\bm K_{\mathrm{LCP}}\bm K_{\mathrm{LCP}}^{\top}$ defines an output ellipsoid.
Hence, $\bm K_{\mathrm{LCP}}$ acts as an anisotropic ``Lipschitz budget'' that shapes the direction-dependent
sensitivity of the policy.
When $\bm{K}_{\mathrm{LCP}}$ is chosen SPD, this can be written as
\begin{equation}
\bm{J}_{\pi}(\bm{o})\,\bm{J}_{\pi}(\bm{o})^{\!\top}
\;\preceq\;
\bm{K}_{\mathrm{LCP}}^{2},
\quad \forall\,\bm{o}\in\mathcal{O}.
\label{eq:hard_matrix_constraint_psd}
\end{equation}

\subsubsection{Spectral (max-eigenvalue) violation}
Define the violation as
\begin{equation}
v_{\max}(\bm{o})
\triangleq
\lambda_{\max}\!\left(
\bm{J}_{\pi}(\bm{o})\,\bm{J}_{\pi}(\bm{o})^{\!\top}
-
\bm{K}_{\mathrm{LCP}}^{2}
\right),
\end{equation}
where $\lambda_{\max}(\cdot)$ represents the maximum eigenvalue;
and penalize only positive violations:
\begin{equation}
\mathcal{R}_{\mathrm{lip}}^{\max}(\bm{\theta})
=
\mathbb{E}_{\bm{o}\sim D}\Big[\big(v_{\max}(\bm{o})\big)_+^2\Big].
\label{eq:matrix_hinge_maxeig}
\end{equation}
{where we define the positive-part operator as
$
x_{+} \triangleq \max(x,0)
$ for any scalar $x\in\mathbb{R}$. 
Based on \eqref{eq:hard_matrix_constraint_psd}, $\bm{J}_{\pi}(\bm{o})\,\bm{J}_{\pi}(\bm{o})^{\!\top}
-
\bm{K}_{\mathrm{LCP}}^{2}$ should always be negative semidefinite. 
Accordingly, $(v_{\max}(\bm o))_{+}^{2}$ penalizes only constraint violations ($v_{\max}(\bm o)>0$) and is zero otherwise.}
This yields the unconstrained objective in the loss function
\begin{equation}
\mathcal{L}_{\mathrm{total}}
=
\mathcal{L}_{\mathrm{RL}}
+
\lambda_{\mathrm{lip}}\,
\mathcal{R}_{\mathrm{lip}}^{\max}(\bm{\theta}).
\label{eq:loss_total_matrix_maxeig}
\end{equation}
where $\lambda_{lip}$ is a weight of LCP loss. 

\subsection{Joint-space Stiffness Upper-bound Ellipsoid}

\label{sec:anisotropic_stiffness_bound}

Let $\bm{x}=f(\bm{q})$ be a task-space variable with Jacobian
$\bm{J}_x(\bm{q})=\partial \bm{x}/\partial \bm{q}\in\mathbb{R}^{m\times n}$.
For clarity, we consider \emph{anisotropic} (matrix-valued) stiffness budgets specified by symmetric positive-definite matrices
$\bm{K}_x^{\max}\succ \bm{0}$.

According to \cite{Yamamoto2017ICRA}, the contact conditions of a humanoid robot can be categorized into single-support and double-support. In each contact mode, the support foot is assumed to be a fixed base (i.e., no relative slipping). 
In the single-support case, the relationship between task-space compliance {$\bm C_x$ and joint-space compliance $\bm C_{q}$} is 
\begin{align}
\label{single_stiff1}
 \bm J_x \bm C_{q} \bm J_x^{\!\top} = \bm C_x .
\end{align}
A general symmetric solution of \eqref{single_stiff1} is
\begin{align}
\label{eq:osc_equivalent_stiff}
\bm{C}_{q}
	=
	\bm{J}_x^{\sharp} \bm{C}_x \bm{J}_x^{\top\sharp}
	+ 
	(\bm{Y} - \bm{J}_x^{\sharp}\bm{J}_x \bm{Y} \bm{J}_x^{\!\top}\bm{J}_x^{\top\sharp} ) ,
\end{align}
where $\bm J_x^{\sharp}$ is the pseudo-inverse of $\bm J_x$, and $\bm Y$ is an arbitrary symmetric matrix for null-space compliance.


Given a task-space stiffness upper bound $\bm K_x^{\max}$, \eqref{eq:task_ellipsoid_constraint_goal} is equivalently expressed as the compliance lower-bound
\begin{align}
\bm C_x \ \succeq\ \bm C_x^{\max}, \\ \bm C_x^{\max}\triangleq (\bm K_x^{\max})^{-1}.
\label{eq:Cx_lower_bound}
\end{align}
Using \eqref{single_stiff1}, a sufficient condition for \eqref{eq:Cx_lower_bound} is to choose a joint-space compliance budget $\bm C_q^{\max}$ such that
\begin{equation}
\bm J_x\,\bm C_q^{\max}\,\bm J_x^{\top} \ \succeq\ \bm C_x^{\max},
\label{eq:compatibility_condition}
\end{equation}
Finally, defining $\bm K_q^{\max}\triangleq (\bm C_q^{\max})^{-1}$, and {$\bm K_{\mathrm{q}}\triangleq \bm C_{q}^{-1}$} as the joint stiffness matrix, we obtain the following sufficient joint-space condition:
\begin{equation}
\bm K_{\mathrm{q}} \ \preceq\ \bm K_q^{\max}
\label{eq:Kq_sufficient}
\end{equation}
which guarantees the task-space bound~\eqref{eq:task_ellipsoid_constraint_goal}.
{In this study}, we consider the joint-space stiffness $\bm K_{\mathrm{q}}$ to be physically equivalent to $\bm K_{\mathrm{eq}}$ {in \eqref{eq:Kq_equiv_prop}}: {$\bm K_{\mathrm{q}} = \bm K_{\mathrm{eq}}$}.

\subsection{Anisotropic Lipschitz-Constrained Policies Based on Target Stiffness \label{sec:stiffness_to_lcp_bound}}


\subsubsection{Anisotropic upper-bound via an ellipsoidal metric induced by $\bm{K}_q^{\max}$}
Let $\bm{K}_q^{\max}\in\mathbb{R}^{n\times n}$ be {an arbitrary} symmetric PSD matrix encoding an anisotropic closed-loop stiffness upper-bound.
Assuming $\bm{K}_q^{\max}\succ \bm{0}$,
we define the weighted norm induced by compliance $\bm{C}_q^{\max} = (\bm{K}_q^{\max})^{-1}$: 
\begin{equation}
\|\bm{v}\|_{\bm{C}_q^{\max}}
\triangleq
\sqrt{\bm{v}^\top(\bm{K}_q^{\max})^{-1}\bm{v}}.
\label{eq:weighted_norm_def}
\end{equation}
We impose the following \emph{anisotropic gain constraint} on torque perturbations:
\begin{equation}
\|\delta\bm{\tau}\|_{\bm{C}_q^{\max}}
\le
\|\delta\bm{q}\|_2,
\quad
\forall\,\delta\bm{q}\in\mathbb{R}^n.
\label{eq:anisotropic_gain_constraint}
\end{equation}
{Here enforcing \eqref{eq:anisotropic_gain_constraint} for all $\delta \bm q$ ensures the closed-loop incremental torque response is bounded in every joint-space direction.}
Substituting \eqref{eq:delta_tau_linearization} into \eqref{eq:anisotropic_gain_constraint} yields
\begin{equation}
\|\bm{K}_p(\bm{I}-\alpha\bm{J}_\pi)\,\delta\bm{q}\|_{\bm{C}_q^{\max}}
\le
\|\delta\bm{q}\|_2,
\quad
\forall\,\delta\bm{q}.
\label{eq:anisotropic_gain_constraint_sub}
\end{equation}

\subsubsection{Reduction to a Lipschitz-style (spectral-norm) constraint\label{sec:silc_def}}
Let $\bm{K}_q^{\max}=\bm{L}\bm{L}^\top$ be a Cholesky factorization with $\bm{L}$ invertible. Then
\begin{equation}
\|\bm{v}\|_{\bm{C}_q^{\max}}
=
\|\bm{L}^{-1}\bm{v}\|_2.
\label{eq:whitened_norm}
\end{equation}
Using~\eqref{eq:whitened_norm}, the constraint~\eqref{eq:anisotropic_gain_constraint_sub} is equivalent to
\begin{equation}
\left\|\bm{L}^{-1}\bm{K}_p(\bm{I}-\alpha\bm{J}_\pi)\right\|_2 \le 1.
\label{eq:routeA_spectral_constraint}
\end{equation}
In this study, we define the constraint in \eqref{eq:routeA_spectral_constraint} as Stiffness-Induced Lipschitz Constrained (SILC). 
Up to this point, we can observe that although SILC also imposes a Lipschitz-type constraint on the policy Jacobian, scalar LCP essentially constrains the Jacobian matrix itself, whereas the SILC constrains the equivalent stiffness directly.
Therefore, tightening scalar LCP can only drive the effective stiffness to asymptotically approach the underlying PD stiffness, whereas SILC can constrain the joint stiffness to lie below an arbitrary target upper bound—even lower than the stiffness of the low-level PD controller. 

\subsubsection{Training objective}
Define the whitened Jacobian-like quantity
\begin{equation}
\bm{A}_{\theta}(\bm{o})
\triangleq
\bm{L}^{-1}\bm{K}_p\!\left(\bm{I}-\alpha\bm{J}_\pi(\bm{o}) \right)
\label{eq:A_theta_def}
\end{equation}
A hinge-squared penalty that softly enforces~\eqref{eq:routeA_spectral_constraint} is
\begin{equation}
\mathcal{R}_{\mathrm{aniso}}(\bm{\theta})
=
\mathbb{E}_{\bm{o}\sim D}
\left[
\big(\|\bm{A}_{\theta}(\bm{o})\|_2-1\big)_+^2
\right],
\label{eq:routeA_penalty}
\end{equation}
where $\|\cdot\|_2$ denotes the spectral norm (largest singular value), which can be approximated efficiently (e.g., via power iteration).
Note that the stiffness upper-bound ellipsoid is not absolutely inviolable, we therefore still adopt a soft-constraint formulation. 
In addition, the spectral-norm penalty used in this context is still a form of gradient penalty rather than an SN-LCP \cite{shin2025spectral}.

The total training loss can be written as
\begin{equation}
\mathcal{L}_{\mathrm{total}}(\bm{\theta})
=
\mathcal{L}_{\mathrm{RL}}(\bm{\theta})
+
\lambda_{\mathrm{aniso}}\,
\mathcal{R}_{\mathrm{aniso}}(\bm{\theta}).
\label{eq:routeA_total_loss}
\end{equation}
where $\lambda_{\mathrm{aniso}}$ is the weights of anisotropic Lipschitz loss.

\section{Reinforcement Learning Setup \label{sect:train}}

\subsection{Baseline}
This section describes the configuration of the RL training in Isaac Gym environment used in this work.
For upper-body motion priors, we employ the AMASS dataset \cite{mahmood2019amass}. 
In the baseline, the network architecture and hyperparameter settings are kept consistent with those of FALCON \cite{zhang2025falcon}. 
Moreover, to enable a more direct comparison with the proposed method, we modify the baseline from the original work as follows.
First, we augment the observation space with the contact phase and upper-body torque. 
Second, we reduce the weight of the regularization (joint velocity, acceleration, and torque) reward term by scaling it by a factor of 0.1.

\subsection{Training Details}
In this work, all training is conducted on a workstation running Ubuntu 22.04, equipped with an Intel Core i9-14900K CPU and an NVIDIA RTX 4080 GPU.
The frequency of the policy output is 50 Hz. The action scale $\alpha$ is 0.25. 
The policy parameter $\bm \theta$ are updated via proximal policy optimization (PPO) \cite{schulman2017proximal}. 
As in the baseline, the observation is constructed by stacking the states from the previous five time steps.
However, when computing the equivalent stiffness, we use only the state from the previous time step to evaluate the policy Jacobian.
For the introduced SILC loss term, its weight $\lambda_{\mathrm{aniso}}$ is set to 0.5.
To select the appropriate kinematic chain for computing the equivalent joint stiffness, we employ a finite-state machine (FSM) during training to determine whether the robot is in double-support, single-support, or a flight phase based on foot-contact detection.
Because the foot contact force is privileged information that is required only during training, it does not pose a concern for deployment, where such measurements may be unavailable.

\section{Experiment \label{sect:simulation}}

\subsection{Experiment Setting \label{sec:exp_set}}
This section primarily evaluates the effect of SILC in both simulation and hardware experiments and compares it against the baseline and scalar LCP.
We mainly compare the following approaches: 
\begin{itemize}
    \item \textbf{Baseline (without LCP)}: similar as in FALCON \cite{zhang2025falcon}, but without action jitter penalty. 
    \item \textbf{Scalar LCP}: set Lipschitz upper-bound as a scalar \cite{shin2025spectral}:
    \begin{align}
        \|\bm{J}_{\boldsymbol{\theta}}(\bm{o})\|^2
\;\leq\;
{K}_{\mathrm{LCP}}^{2},
\quad \forall\,\bm{o}\in\mathcal{O}.
    \end{align}
    We set the Lipschitz constant $K_{LCP}$ values to $2.0$ and weight $\lambda = 0.5$.
    Empirically, this Lipschitz constant is already close to the minimum critical threshold; reducing $K_{LCP}$ further would directly lead to control instability.
    \item \textbf{Task stiffness-induced anisotropic LCP}: in this case, we set a task stiffness-induced Lipschitz upper-bound in the policy. 
    We configured three modes, namely \textit{high-stiff}, \textit{compliance}, and \textit{hard-nullspace}. 
    The specific stiffness settings are provided in \tabref{tab:1}. 
    $\lambda_{\mathrm{aniso}}$ is set to 0.5. 
\end{itemize}

\begin{table}[t]
\centering
\caption{CoM task and null-space stiffness setting in SILC method. \label{tab:1}}
\begin{tabular}{c|c|c}
\toprule
\textbf{Mode}& \textbf{CoM Stiffness} & \textbf{Null-space Stiffness} \\
\hline
\textbf{High-stiff} & $[1000, 1000, 20000]$ & $100.0$ \\
\textbf{Compliance} & $[200, 200, 20000]$ & $50.0$ \\
\textbf{Hard-nullspace} & $[200, 200, 20000]$ & $500.0$ \\
\bottomrule
\end{tabular}
\end{table}

To evaluate the performance of each approach, several metrics are used as follows. 
Beyond several commonly used evaluation criteria (e.g., joint torque, joint velocity), this paper follows \cite{chen2025learning} and introduces the following metric definitions:
\begin{enumerate}
    \item \emph{Jitter} (action, DoF velocity, torque): 
Let $\bm a_t \in \mathbb{R}^n$ be the action vector at discrete time $t$,
sampled with constant timestep $\Delta t$:
\begin{align}
\label{eq:jitter}
\frac{1}{T-3}\sum_{t=4}^{T}
\frac{1}{n}\sum_{i=1}^{n}
\left|
\frac{a_{t,i}-3a_{t-1,i}+3a_{t-2,i}-a_{t-3,i}}{\Delta t^{3}}
\right| .
\end{align}
To compute other jitters, one can simply replace $a_{\star}$ in \eqref{eq:jitter} with $\dot{q}_{\star}$ or $\tau_{\star}$, respectively.
    \item \emph{Energy} (absolute-mean):
\begin{equation}
\frac{1}{T}\sum_{t=1}^{T}
\frac{1}{n}\sum_{i=1}^{n}
\left|
\tau_{t,i}\,\dot{q}_{t,i}
\right| .
\end{equation}

\item \emph{Directional Quadratic-form Budget}: 
We enforce the stiffness upper bound of \eqref{eq:task_ellipsoid_constraint_goal} in a directional (quadratic-form) sense for any joint perturbation $\delta\bm{q}\in\mathbb{R}^n$:
\begin{equation}
\delta\bm{q}^{\top}\bm{K}_{\mathrm{eq}}(\bm{o})\,\delta\bm{q}
\ \le\
\delta\bm{q}^{\top}\bm{K}_{\mathrm{eq}}^{\max}\,\delta\bm{q}.
\label{eq:Keq_directional_ineq}
\end{equation}
That is, the incremental elastic energy induced by $\bm{K}_{\mathrm{eq}}(\bm{o})$ is upper bounded by the budget ellipsoid defined by $\bm{K}_{\mathrm{q}}^{\max}$ in every direction $\delta\bm{q}$.

\item \emph{High-frequency energy}:
\begin{equation}
\sum_{i\in\{x,y,z\}}
\int_{f_0}^{f_N} S_{x_i x_i}(f)\,df,
\end{equation}
where $S_{x_i x_i}(f)$ is the power spectral density of $i$-th task position component, $f_0$ is 5 Hz, and $f_N$ is 100 Hz.

\item \emph{Peak-to-peak (P95-P5)}:
\begin{equation}
\operatorname{P}_{95}\!\big(s(t)\big)
-
\operatorname{P}_{5}\!\big(s(t)\big),
\end{equation}
where $s(t)$ denote a scalar trajectory signal of interest, and $\operatorname{P}_{p}(\cdot)$ denotes the $p$-th percentile over the evaluation interval.

\end{enumerate}




\subsection{Simulation Results}

In this subsection, all simulations are conducted on robot Unitree G1 in the Isaac Gym environment.


  

\subsubsection{CoM Compliance in Stance Mode}

\begin{figure}[t]
	\begin{center}
	 \includegraphics[width=0.9\hsize]{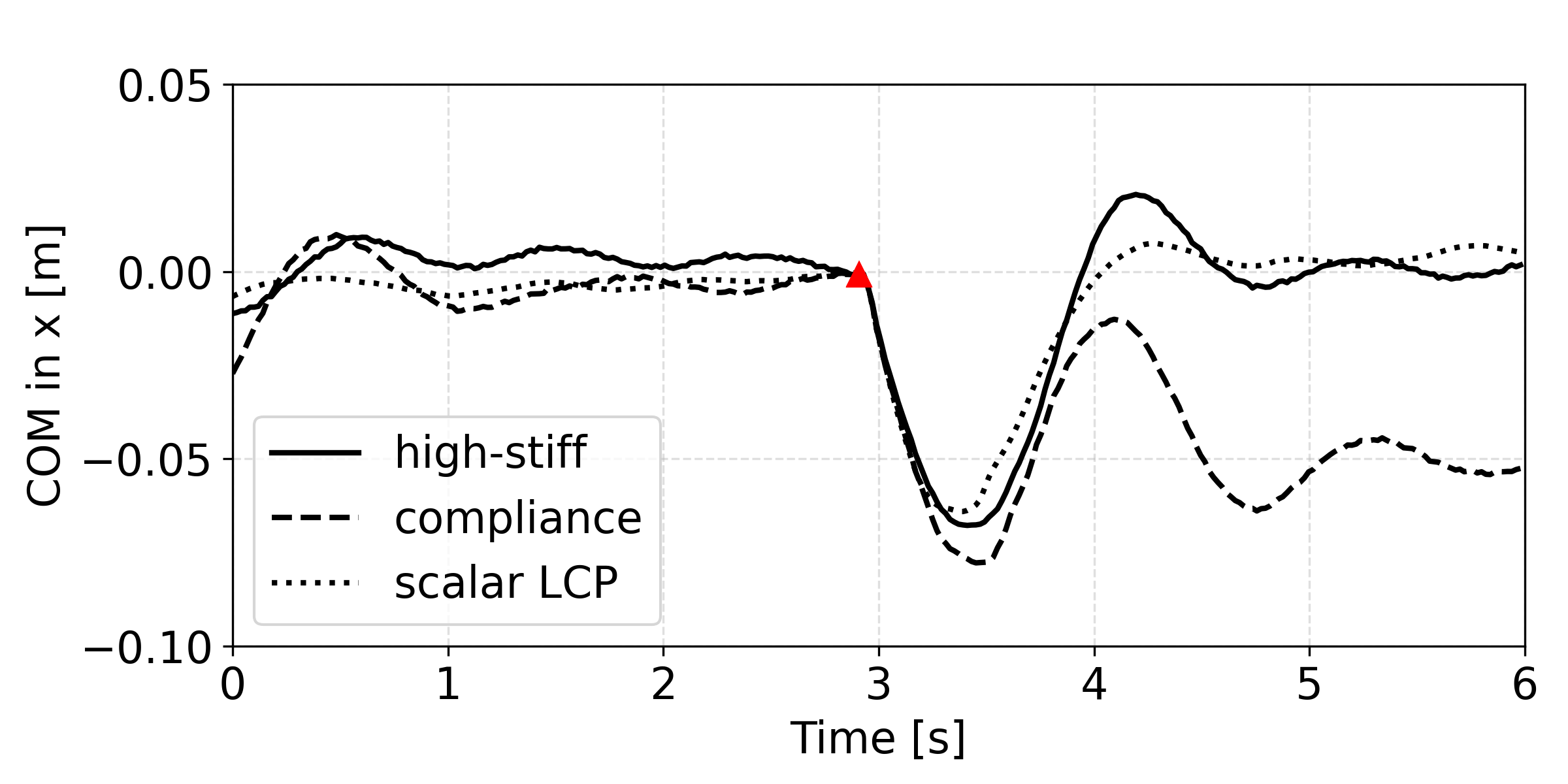}
	 \caption{The $x$-direction CoM trajectory receiving an $-x$-direction impact (150 N in 0.05 s) on pelvis link. The red triangle indicates the time instant at which the impact is applied. Compared with the \textit{high-stiff} mode—in which the CoM returns to its original state via a damped oscillatory response—the \textit{compliance} mode exhibits more pronounced compliance, as evidenced by larger oscillation amplitudes and the lack of convergence back to the original state. Meanwhile, \textit{scalar LCP} also exhibits similarly high stiffness to high-stiff mode.
}
    \label{fig:com_impact}
	\end{center}
\end{figure}

In the double-support standing mode, we first enable only the CoM compliance task.
In this experiment, we apply an external impulse (-150 N in 0.05 s) to G1's pelvis link in $x$-direction while it is standing.
\figref{fig:com_impact} illustrates the CoM fluctuation trajectories under the compliance SILC and high-stiff SILC settings.
It can be clearly seen from the figure that, under the same-magnitude impact, the CoM in the compliance case deviates and then returns to the initial equilibrium state. In contrast, in the high-stiff case, the CoM exhibits a larger deviation and does not return to its original position. 
We attribute this to the larger CoM compliance in the compliance case: the external impact induces a larger displacement, and subsequently, the policy prioritizing balance recovery sacrifices CoM tracking accuracy.
This result indicates that, under the proposed SILC method, specifying different task-space stiffness constraints can induce distinct compliance behaviors in the robot.
We also observe that the response behavior of scalar LCP is similar to that of high-stiff case.
As discussed earlier, this Lipschitz constant is already the smallest value that preserves stability; however, it still cannot satisfy the requirement of high CoM compliance. This is mainly because the two kind of LCP are fundamentally different in essence as mentioned in \sectref{sec:silc_def}.

\subsubsection{End-effector Compliance Under Constant Load\label{sec:hand_load}}

\begin{figure}[t]
	\begin{center}
	 \includegraphics[width=0.9\hsize]{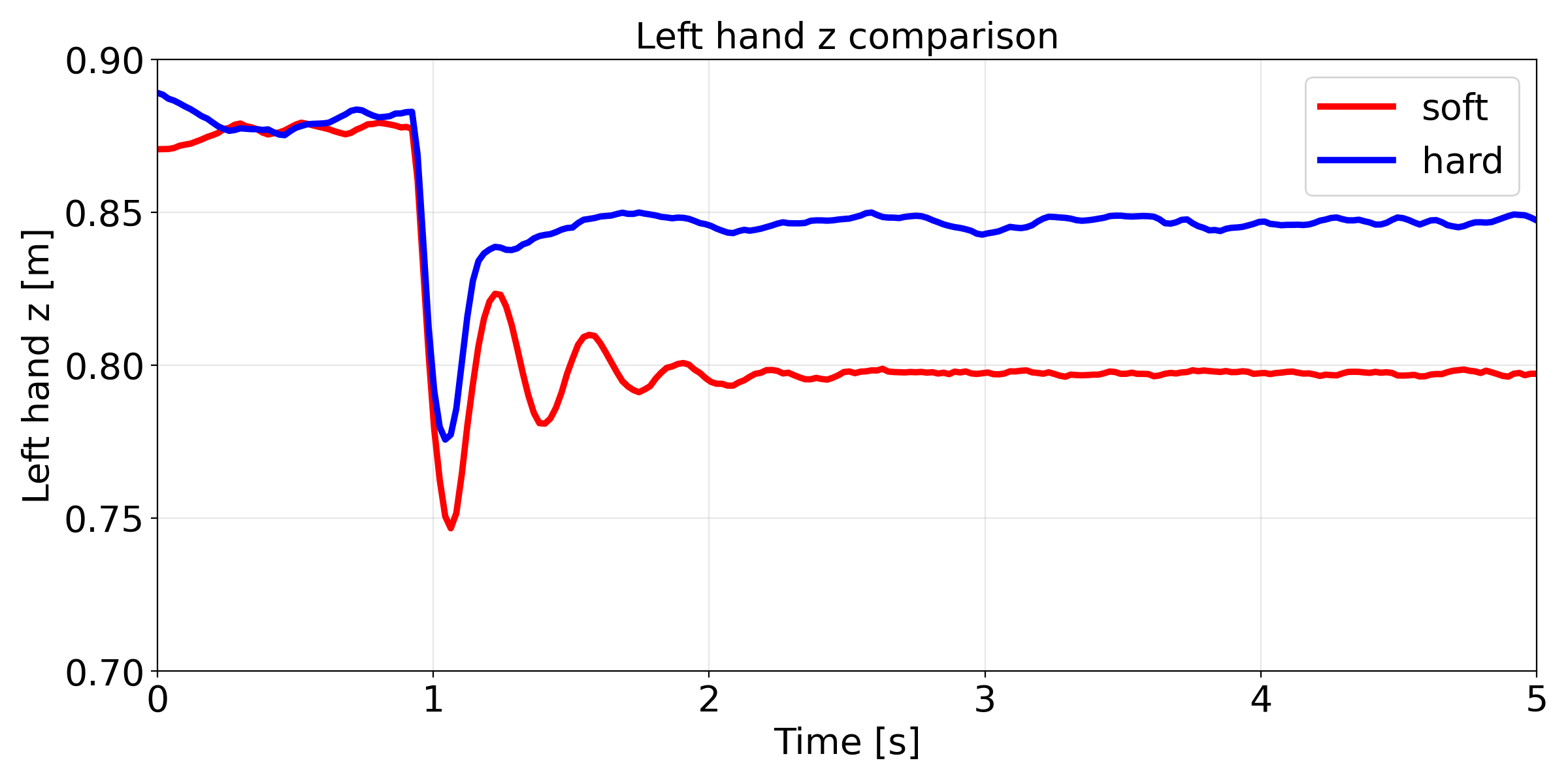}
	 \caption{Hand displacement under different task-space compliance settings of SILC when a constant load (-40 N) is applied to the left hand. 
     It can be observed that, in both cases, the hand displacement responds within a very short time ($<$ 0.1 s). 
    After the response settles, we compare the steady-state position with the initial position and, together with the applied external force, approximately estimate the equivalent task-space stiffness to be 400 (soft) and 1000 N/m (hard), respectively.
}
    \label{fig:left_hand}
	\end{center}
\end{figure}

In this subsection, the G1 robot stands in double-support while its arms are set to maintain a default L-shape. 
A constant vertical external force (-40 N in $z$-axis) is applied to the robot’s left hand.
We compared two cases both applying SILC but using different parameter settings. 
In both cases, CoM and null-space stiffness are the same as in the compliance case in \tabref{tab:1}.
While in the \textit{hard} case, the hand task stiffness is $[1000, 1000, 1000]$; in the \textit{soft} case, hand stiffness is $[200, 200, 200]$. 
\figref{fig:left_hand} illustrates the $z$-direction displacement of the left hand during the application of the external force. 
By combining the displacement between the quasi-steady-state position and the initial position with the known external force, we estimate the measured equivalent task-space stiffness under the two control policies to be approximately 400 (compliance) and 1000 N/m (high-stiff), respectively.
Compared with the setting compliance, this result demonstrates that the proposed SILC not only enforces stiffness constraints in fast transient responses, but also enables controllable compliance under quasi-static deformations induced by external forces.

\subsubsection{Walking Motion}

\begin{figure}[t]
  \centering
  \subfloat[Proposed compliance SILC]{
    \includegraphics[width=.85\linewidth]{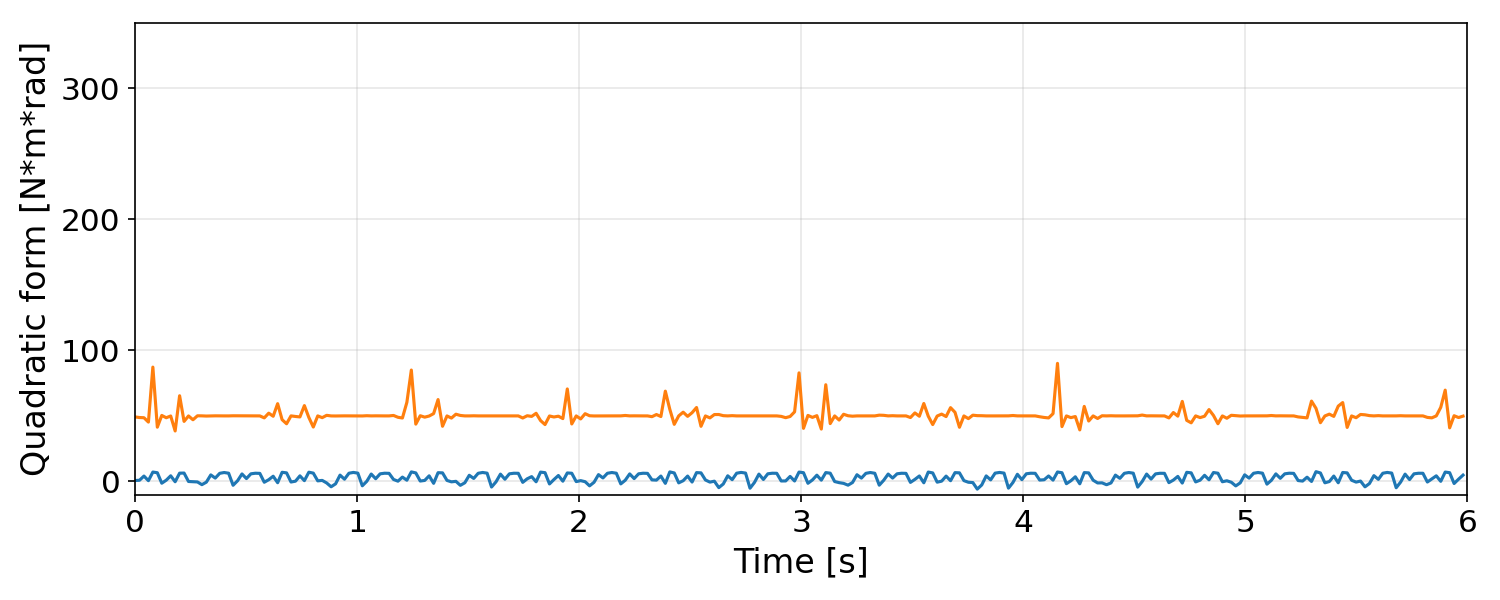}
  }
  
  \subfloat[Scalar LCP]{
    \includegraphics[width=.85\linewidth]{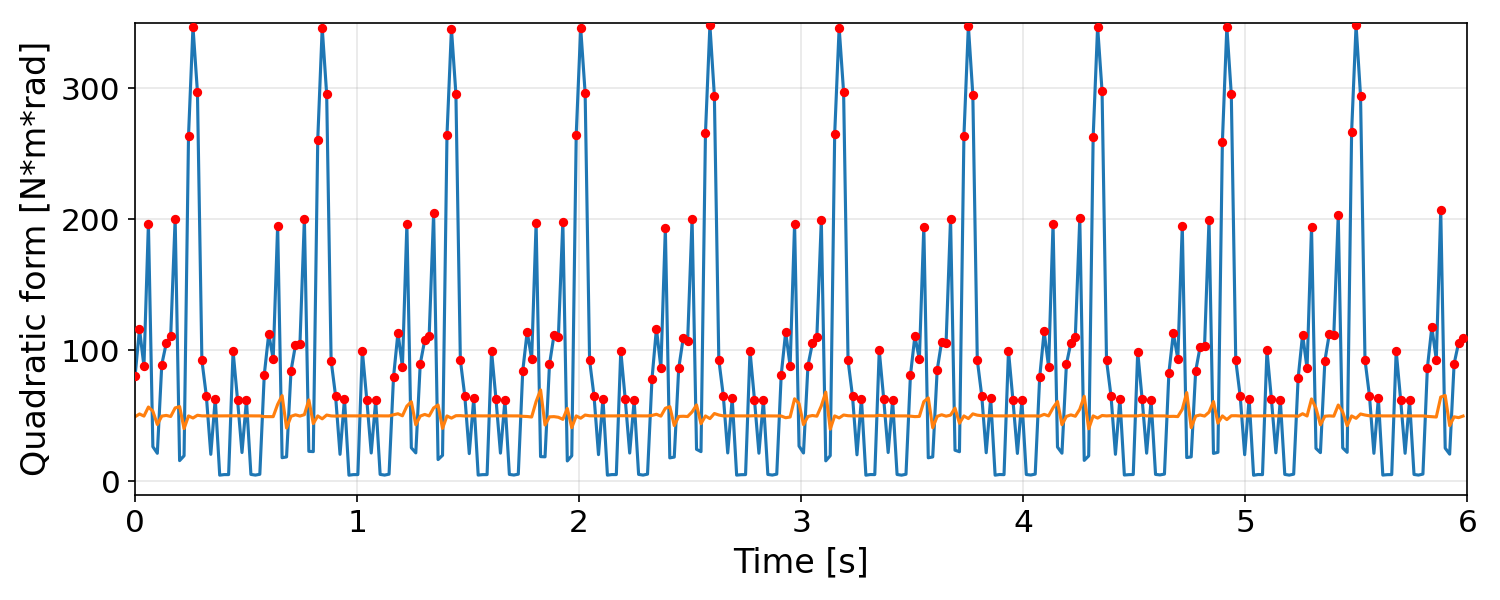}
  }

  \subfloat[Baseline]{
    \includegraphics[width=.85\linewidth]{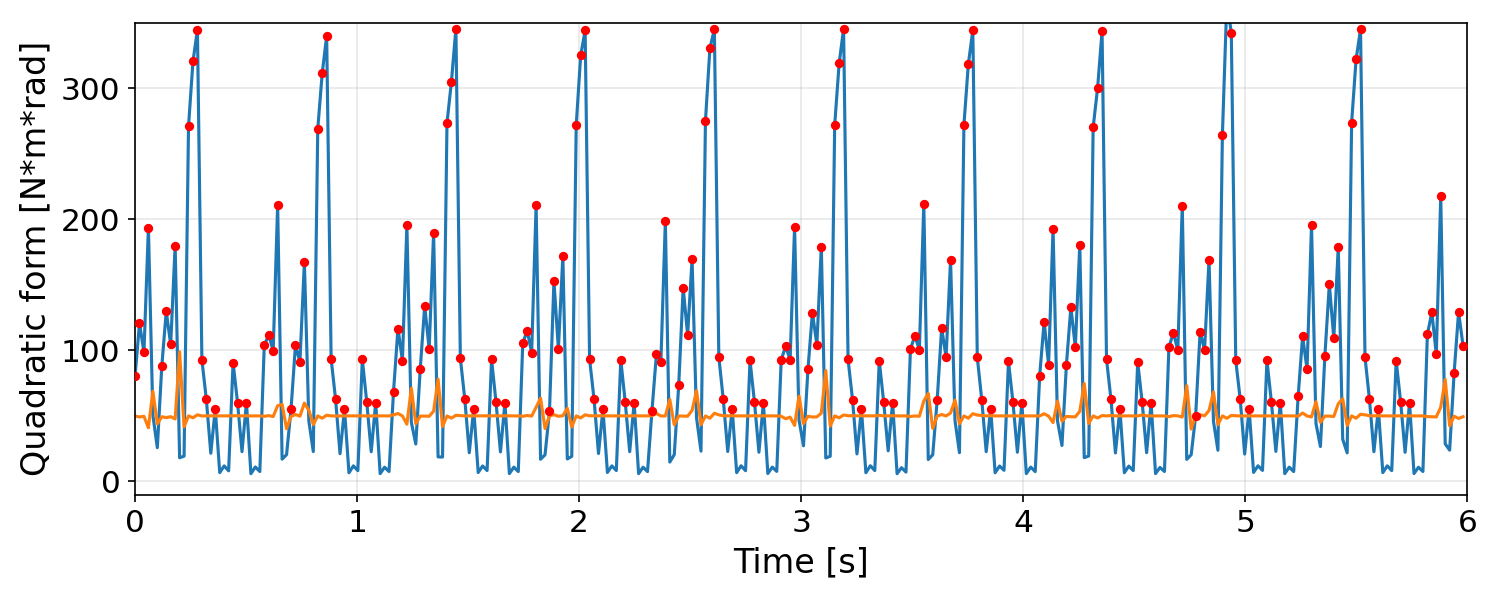}
  }
  \caption{According to the definition of the Directional Quadratic-form Budget, at each control step we plot the left-hand side (LHS) and right-hand side (RHS) of \eqref{eq:Keq_directional_ineq}, respectively (unit of quadratic form: N$\cdot $m$\cdot $rad).
  The yellow curve corresponds to the RHS of the equation, i.e., the quadratic form under the target $\bm{K}_{q}^{\max}$, whereas the blue curve corresponds to the LHS, i.e., the current quadratic form measured $\bm{K}_{q}(\bm o)$. 
  The red markers indicate instances where the measured value exceeds the target. As shown, under the SILC method, the measured value is strictly bounded by the target constraint; in contrast, under scalar LCP and baseline, the measured quadratic form periodically exceeds the target during the stepping gait.
}
  \label{fig:kq_budget}
\end{figure}

In this subsection, the robot switches between different supporting-cases by the FSM. 
We first evaluate the violations of the directional PSD budget as in \sectref{sec:exp_set} during a stepping-in-place motion while the upper limbs track a random trajectory.
In this evaluation, we randomly sample unit $\delta \bm q$ vectors along different DoFs to compute the PSD values as in \eqref{eq:Keq_directional_ineq}.
\figref{fig:kq_budget} illustrates the left-hand side and right-hand side values of \eqref{eq:Keq_directional_ineq}. 
We can find that the measured value under SILC is strictly bounded by the budget; while under scalar LCP and baseline, the measured quadratic form periodically exceeds the calculated budget.

We then sample random velocity commands for the robot across 1,024 environments. 
First, after training for a given number of iterations (14,000), the basic tracking rewards are very similar across all settings; we therefore treat this as a fair baseline for comparison.
Then we measure the corresponding metrics over an episode, as summarized in \tabref{tab:2}.
The comparison of all LCP variants with the baseline shows that the LCP methods yield noticeably smaller energy values and jitter of action and torque, but a larger velocity jitter.
We attribute this primarily to the fact that the LCP constrains the output stiffness at the policy-Jacobian level, thereby substantially suppressing the baseline’s tendency to achieve task tracking via high-frequency oscillations.
Next, relative to the scalar LCP baseline, SILC methods reduce action jitter but increase energy consumption. 
By decomposing the metrics into upper- and lower-body, we find that this energy increase is driven primarily by the lower body, which also exhibits higher jitter under SILC.
These results suggest that SILC attains compliance by reallocating effort to the lower body, partially trading motion smoothness for compliance—contrary to the common assumption that greater compliance lowers energy. 
Notably, SILC compliance is task-conditioned on simultaneously maintaining balance and tracking upper-body motion, making this trade-off reasonable. 
Consistently, a higher stiffness setting (high-stiff) can even reduce energy consumption.

\begin{table*}[t]
  \centering
  \caption{Comparison of walking motion under random velocity command. (environment number: 1,024) \label{tab:2}}
  \begin{tabular}{lcccccc}
    \toprule
     & \textbf{Action Jitter} ($\times 10^4$) $\downarrow$ & \textbf{DoF Velocity Jitter [rad/s]} ($\times 10^5$) $\downarrow$ & \textbf{Torque Jitter [Nm]} ($\times 10^5$) $\downarrow$ & \textbf{Energy [J]} $\downarrow$ \\
    \midrule
    \textbf{High-stiff SILC}    & $3.08 \pm 0.19$ & $1.97 \pm 0.17$ & $4.89 \pm 0.34$ & $\bm{4.77 \pm 3.07}$ \\
    \textbf{Compliance SILC}    & $3.19 \pm 0.28$ & $2.03 \pm 0.28$  & $5.04 \pm 0.38$ & $4.99 \pm 5.17$ \\
    \textbf{Hard-nullspace SILC}    & $\bm{3.03 \pm 0.17}$ & $1.84 \pm 0.16$  & $4.80 \pm 0.22$ & $4.86 \pm 4.47$ \\
    \hline
    \textbf{Scalar LCP ($\lambda = 0.5$)}    & $3.52 \pm 7.00$ & $1.74 \pm 1.96$ & $\bm{4.47 \pm 5.01}$ & $5.09 \pm 4.76$ \\
    \hline
    \textbf{Baseline}  & $4.13 \pm 1.94$ & $\bm{1.68 \pm 0.51}$  & $5.35 \pm 1.13$ & $7.58 \pm 10.74$ \\
    \bottomrule
  \end{tabular}
\end{table*}
\subsection{Experiment Validation}
In this subsection, we deploy the previously trained policy on the physical G1 robot. 
At first, we evaluate the hand compliance while the robot is standing, as shown in \figref{fig:g1_exp_a}.
We then evaluate the control performance under a hand load. 
Specifically, we attach a 2.5 kg weight to the robot’s right hand using a rope and release it from a certain height (as shown \figref{fig:g1_exp_b}).
The robot’s right hand is displaced by the external force and oscillates, and then converges to an equilibrium position. 
\tabref{tab:3} summarizes the hand oscillation behavior under different policies.
In the SILC situation, we applied the policy trained by SILC with the \textit{soft} hand task setting as introduced in \sectref{sec:hand_load}. 
As shown in the table, the settling time under SILC is about half that of the baseline, and the high-frequency energy along the $z$-axis is nearly an order of magnitude lower.
This indicates that the proposed SILC method effectively attenuates the oscillations induced by external impacts by increasing hand compliance.




We also similarly set up a stepping motion and placed soft pads on the ground to emulate uneven terrain (as shown in \figref{fig:g1_exp_c}).
During stepping, we recorded the IMU measurements, as shown in \tabref{tab:4}.
As shown in the table, the SILC method (compliance setting) outperforms the baseline across all metrics, for both the RMS (root mean square) and peak-to-peak values.
In particular, the gyroscope fluctuations under SILC are roughly half of those observed with the baseline. 
This result indicates that, compared with the baseline, the proposed SILC method reduces body-link oscillations during locomotion by increasing compliance.

\begin{table}[t]
  \centering
  \caption{Metrics of the right hand $z$-direction trajectory oscillations under external impacts.\label{tab:3}}
  \begin{tabular}{l|cc}
    \toprule
     & \textbf{Settling time [s]} $\downarrow$ & \textbf{High-frequency energy [J]} $\downarrow$ \\
    \hline
    \textbf{SILC (soft)}  & $1.2$ & $1.01 \times 10^{-8}$ \\
    \textbf{Baseline} & $2.4$ & $10.0 \times 10^{-8}$ \\
    \bottomrule
  \end{tabular}
\end{table}



\begin{table}[t]
  \centering
  \caption{IMU measurement in the stepping experiment. \label{tab:4}}
  \begin{tabular}{c|ccc}
    \toprule
     & & \textbf{RMS} $\downarrow$ & \textbf{Peak-to-Peak (P95-P5)} $\downarrow$ \\
    \hline
    \textbf{Compliance}  & Gyro. & $0.247$ & $0.726$  \\
    \textbf{SILC}  & Acc. ($x$ and $y$) & $3.885$ & $12.66$  \\
    \hline
    \textbf{Baseline} & Gyro. & $0.472$ & $1.433$ \\
     & Acc. ($x$ and $y$) & $4.859$ & $13.80$ \\
    \bottomrule
  \end{tabular}
\end{table}

\subsection{Discussion}

Combining the above theoretical derivations and the experimental results, imposing an ALCP on the state-dependent equivalent joint stiffness is typically considered to yield the following effects:


\begin{enumerate}

\item \textbf{Equivalent compliance observer in RL framework.}
Through the proposed policy-induced equivalent joint stiffness, the stiffness under a RL controller can be explicitly observed. This equivalent stiffness is neither simply the stiffness of the low-level controller nor the stiffness estimated from a task-space displacement–torque relationship; rather, it is the true effective stiffness derived from the policy’s inherent characteristics.

\item \textbf{Adjustable compliance and impact robustness.}
In the proposed SILC method, the equivalent joint-stiffness constraint is derived directly from the task-space specification. 
Therefore, task compliance budget can be tuned via the LCP formulation.
Moreover, the experimental results confirm that both the transient and steady-state responses can be modulated using this approach, thereby enabling different robustness under external disturbances.

\item \textbf{Smoother and more stable motion.}
Since the Lipschitz constraint limits the equivalent stiffness gain, the policy cannot react to small observation perturbations with arbitrarily large changes in action. 
In practical control, this property can mitigate the high-frequency jitter induced by the policy, improve motion stability, and also contribute to energy efficiency.

\end{enumerate}

This work has three main limitations. First, although SILC updates its kinematic terms online, the task-space stiffness budget is not yet adjustable at runtime. 
Second, the current formulation constrains only the policy sensitivity to joint positions, so the induced equivalent stiffness is defined on a limited subspace and ignores other relevant state components (e.g., joint velocities, task-space states, or latent variables). 
Finally, we regulate stiffness only; the associated damping design and constraints are left for future work.

\section{Conclusion}

This study proposes a Stiffness-Induced Lipschitz Constrained (SILC) policy that transfers a task-specified stiffness upper-bound to an anisotropic Lipschitz budget on the policy Jacobian and enforces it via a Lipschitz constraint during training. 
Our method builds upon Lipschitz-constrained policies (LCP), but extends the original scalar constraint to an anisotropic form and links it to task-specified compliance.
With the proposed theoretical formulation, compliance during policy execution can be achieved by directly specifying an upper-bound on the task-space stiffness during RL training.







\bibliographystyle{IEEEtran.bst}
\bibliography{ral_hrvc}


\end{document}